\documentclass[conference]{IEEEtran}
\IEEEoverridecommandlockouts
\usepackage{cite}
\usepackage{booktabs}
\usepackage[table]{xcolor}
\usepackage{threeparttable}

\usepackage{amsmath,amssymb,amsfonts}
\usepackage{algorithmic}
\usepackage{graphicx}
\usepackage{textcomp}
\usepackage{xcolor}
\usepackage{url}
\usepackage{xurl}
\usepackage{hyperref}
\usepackage{booktabs}
\usepackage{colortbl}
\usepackage[inline]{enumitem}
\usepackage{ragged2e} 
\usepackage{tabularx}

\def\BibTeX{{\rm B\kern-.05em{\sc i\kern-.025em b}\kern-.08em
    T\kern-.1667em\lower.7ex\hbox{E}\kern-.125emX}}
\begin{document}

\title{LuxEmo: Expressive Text-to-Speech Corpus for Luxembourgish}
\author{
\IEEEauthorblockN{1\textsuperscript{st} Nina Hosseini-Kivanani}
\IEEEauthorblockA{
\textit{Radio T{\'e}l{\'e}visioun L{\"e}tzebuerg (RTL)} \& \textit{University of Luxembourg}\\
Esch-sur-Alzette, Luxembourg\\
ORCID: 0000-0002-0821-9125
}
\and
\IEEEauthorblockN{2\textsuperscript{nd} Sandipana Dowerah}
\IEEEauthorblockA{
\textit{Tallinn University of Technology}\\
Tallinn, Estonia\\
ORCID: 0000-0002-1559-4505
}
}

\maketitle

\begin{abstract}


State-of-the-art speech datasets predominantly focus on widely spoken languages, often overlooking low-resource languages such as Luxembourgish, which remain underrepresented in speech technology research. In this work, we introduce LuxEmo, a 21-hour conversational expressive speech corpus for Luxembourgish with 4 emotion categories. LuxEmo is derived from Radio Télévision Luxembourg (RTL) youth broadcasts, using automated detection followed by human validation. We propose a semi-automatic curation workflow that combines voice activity detection, denoising, language identification, LuxASR-based segmentation, automatic emotion prediction, lexical cues, and targeted human review. Additionally, we benchmark five expressive TTS systems covering German-based cross-lingual transfer, multilingual Luxembourgish support, Luxembourgish adaptation, and non-parametric prosody transfer. Performance is evaluated using both objective metrics and human evaluation.
\end{abstract}


\section{Introduction}
Expressive text-to-speech (TTS) supports conversational agents, media production, assistive technologies, and educational platforms. Emotional prosody, including pitch, rhythm, and intensity, shapes naturalness, empathy, and engagement in synthetic voices~\cite{efthymiou2023empathy,aggarwal2023towards,liu2023emotionally}. Prior work shows that modeling affective cues and user state can foster trust, presence, and prosocial behavior in human-computer interaction~\cite{kosgi2022empathic,efthymiou2023empathy}. Current expressive TTS systems are largely developed and evaluated for English and other high-resource languages, often using acted speech recorded in controlled studio conditions~\cite{mitchell2015value,morrison2020controllable}. Real-world expressive speech, however, involves informal registers, noisy recordings, overlapping speech, code-switching, and naturally imbalanced emotion distributions. These challenges are amplified for low-resource languages, where annotated speech data, speaker coverage, transcriptions, and language-specific TTS tools are limited. Multilingual TTS provides partial coverage, but emotional control is still often evaluated through high-resource surrogate languages or transfer-learning strategies~\cite{Tits2019ExploringTL,Huybrechts2020LowResourceET,latore,Tu2019EndtoendTF}. As a result, expressive corpora for low-resource languages remain scarce, especially for spontaneous speech in realistic media conditions~\cite{latif2021survey,schuller2018speech}. Luxembourgish, the national language of Luxembourg, is technologically low-resource~\cite{gilles2023asrlux}. Existing work has addressed foundational tasks such as ASR and lexical or phonological description~\cite{gilles2023asrlux,nayak2023improving,song2023letz}. Luxembourgish is embedded in a multilingual ecology involving German, French, and English~\cite{govlu2018_language_policy,gilles2013luxembourgish}, and everyday speech frequently includes code-switching~\cite {redinger2010language}. Youth-oriented media adds further challenges, including background music, sound effects, overlapping talk, rapid speaker changes, and informal discourse~\cite{tas20284_youth_digital_luxembourg}. Luxembourgish, therefore, provides a relevant but difficult case for studying expressive TTS beyond clean studio corpora.

\begin{figure*}[!t]  
  \centering  \includegraphics[width=\linewidth]{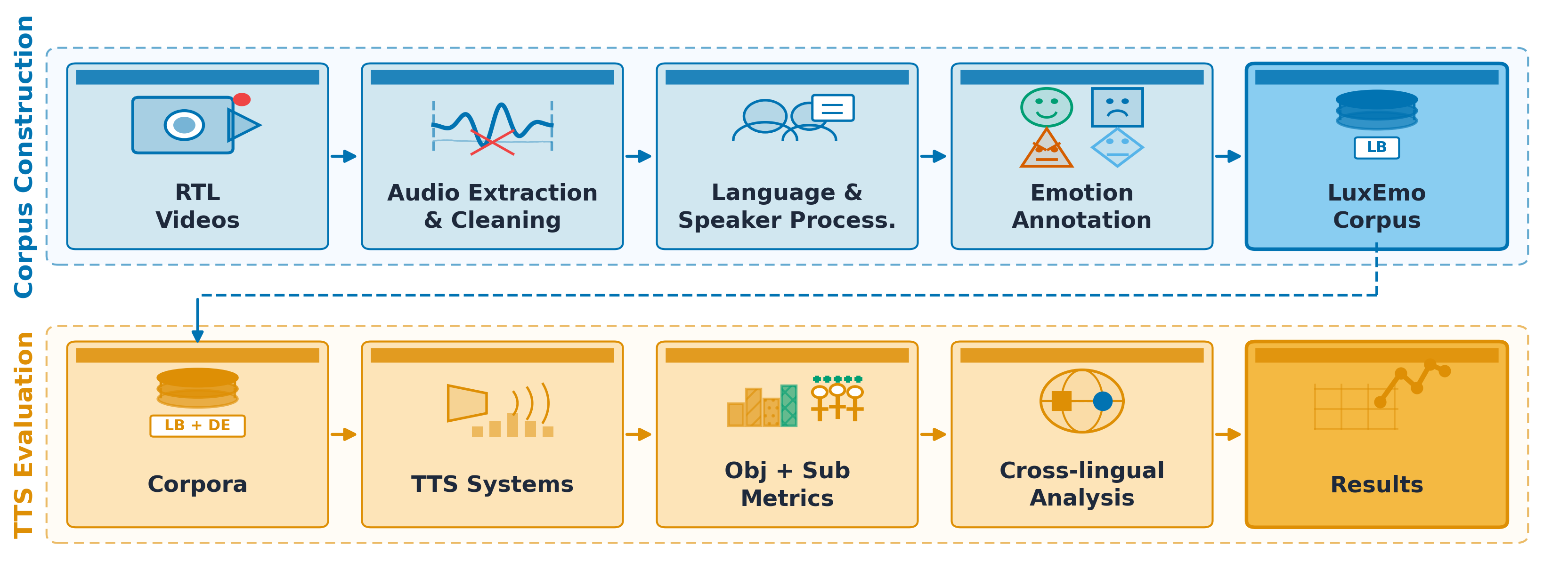}
  \caption{Overall LuxEmo pipeline. Top: corpus construction. 
  Bottom: TTS evaluation pipeline from the corpora.}
  \label{fig:luxemo-overall}
\end{figure*}

This work is resource- and protocol-oriented rather than architectural, where we address the construction and evaluation of an expressive speech corpus for Luxembourgish from spontaneous, code-switching RTL Youth broadcasts. The source material lacks the controlled properties of studio corpora: it contains residual music and noise, overlapping speech, uneven speaker participation, incomplete manual transcripts, and skewed emotion classes. LuxEmo therefore targets a realistic low-resource setting in which clean recordings, balanced labels, and broad speaker coverage are not available. 

We introduce \textit{LuxEmo}, a spontaneous emotional speech corpus and expressive TTS benchmark for Luxembourgish. \textsc{LuxEmo} contains 21 hours of wide-band speech and 7{,}562 segments of up to 10 seconds from four recurring speakers, annotated for language, speaker identity, and four emotion categories in a heavily code-switching setting. First, we document a semi-automatic curation workflow combining voice activity detection, denoising, language identification, LuxASR-based speaker-aware segmentation~\cite{gilles2023asrlux,gilles2023lux}, acoustic and lexical emotion cues, targeted human review, and non-intrusive corpus-quality assessment. The emotion annotations are released as weakly supervised labels with confidence information, not as independently validated gold-standard labels. Second, we provide standardized corpus indices and speaker- and episode-aware train, development, and test splits to reduce leakage from shared conversational context while preserving emotion coverage. Third, we benchmark five TTS systems, including zero-shot multilingual, German-proxy, Luxembourgish-aware, prompt-controlled, and kNN-based prosody-transfer systems. The benchmark characterizes realistic system behavior under heterogeneous deployment conditions. It is not a controlled architectural ablation and should not be interpreted as evidence for a universally best TTS model. LuxEmo complements existing monolingual emotional corpora by offering a testbed for expressive TTS in a low-resource, code-switching setting and by documenting curation strategies for broadcast archives under practical reuse constraints.



\section{LuxEmo Corpus}
\label{sec:corpus}
In this section, we explain the steps involved in creating the LuxEmo dataset. Figure \ref{fig:luxemo-overall} gives an overall graphical description of LuxEmo construction and evaluation.

\subsection{Audio pre-processing}
For the creation of LuxEmo, we collect the data from RTL Youth video programs. RTL programs mainly target teenagers and young adults, combining studio discussions, interviews, and on-location reports. The shows are predominantly in Luxembourgish but include systematic code-switching into German, French, and English. Episodes typically contain multiple recurring hosts and guests, background music, sound effects, and an informal register.

The RTL Youth material consists of approximately 21 hours of spontaneous conversational speech on everyday topics, including multi-speaker interactions and single-speaker segments. From raw broadcast video, we first extract and resample audio to a wideband format suitable for TTS. The resulting signals are pre-processed using Voice Activity Detection (VAD) to remove non-speech segments. We discard segments shorter than 200 ms to remove non-informative speech fragments. The resulting segmentation yields utterances with a mean duration of 9.87 seconds, with $98.5\%$ of segments exceeding 6 seconds, preserving prosodic context while remaining compatible with standard TTS back-ends.

We use DeepFilterNet~\cite{schroter2022deepfilternet} to attenuate background music and ambient noise. The denoised speech files are segmented into fixed-length chunks of 1--10 s and saved as mono 22.05 kHz WAV for downstream TTS. Because ground-truth transcripts were unavailable, segment-level transcriptions were obtained from LuxASR TextGrid files and temporally aligned with VAD-based segments. Single-speaker segments are extracted from LuxASR diarization output and mapped to four recurring speakers. We retain the program-level metadata, such as show identity and approximate air date, to enable later analysis across domains and time.

\paragraph{Quality of preprocessed recordings.}
To characterize the acoustic quality of the processed corpus, we applied NISQA v2.0~\cite{mittag2021nisqa} and DNSMOS~\cite{reddy2021dnsmos} to all 7{,}562 segments after denoising and VAD-based segmentation. NISQA-TTS was not used for this corpus-level analysis because it is calibrated for synthetic speech. The processed corpus obtains a NISQA quality score of 3.38 and a DNSMOS OVRL score of 2.91, consistent with denoised broadcast-origin speech that still contains residual music and ambient noise. The DNSMOS BAK score of 3.73 indicates relatively strong background-noise suppression after preprocessing. These corpus-level values provide an intrinsic quality reference for the synthesized-output analysis in Section~\ref{sec:quality}.

\subsection{Annotation process}

We employed the Wav2Vec2-based \texttt{mms-lid-4017} model~\cite{pratap2024scaling} to do the utterance-level language identification (LID). The model assigns a language label to each segmented utterance. Across 7,562 segments, $83.8\%$ are classified as Luxembourgish, $6.5\%$ as English, and $2.2\%$ as German, while the remaining $7.5\%$ are assigned minority language codes. These cases likely reflect classifier uncertainty, particularly in instances of code-switching, dialectal variation, or non-standard speech patterns. To improve annotation reliability, a subset of 1,135 segments (approximately $15\%$ of the corpus) was manually reviewed by a native speaker of Luxembourgish. This review focused on validating language assignments and identifying potential errors in automatic predictions, particularly in ambiguous or low-confidence segments.

Based on this process, high-confidence items are treated as a supervised subset and used for training and evaluation, while lower-confidence segments are retained for representation learning and weakly supervised experimentation. This design allows the dataset to support both fully supervised and semi-supervised research settings. Gender labels are assigned following speaker identity and are used for descriptive statistics and stratified evaluation. 

\subsection{Emotion detection}

We classified all audio samples into 4 emotion categories, namely, neutral, happy, sad, and angry. Initial labels come from a HuBERT-based classifier applied to each segment, and a lexical model over subtitles or transcripts flags affective keywords and discourse markers. A manually annotated subset is used to tune decision thresholds and refine class definitions, so each segment receives an emotion label with an associated confidence score. High-confidence items form the supervised subset used in training and evaluation, while lower-confidence segments remain available for representation learning and weakly supervised experiments.

\subsection{Data splits}

LuxEmo is partitioned into training, development, and test sets using a speaker-aware and episode-aware splitting strategy. Specifically, all segments originating from a given speaker within the same episode are assigned to the same split. This prevents leakage arising from near-duplicate content or shared conversational context across partitions. To ensure reproducibility, all splits are generated using a fixed random seed (42). The splitting procedure is further constrained to approximate the global emotion distribution while guaranteeing at least one development example for each emotion category and full coverage of all four emotion classes in the test set.

\begin{table*}[!t]
    \centering
    \scriptsize
    \caption{LuxEmo speaker-level statistics and corpus-level comparison with EmoDB.}
    \label{tab:luxemo_stats}

    \begin{threeparttable}
    \setlength{\tabcolsep}{20pt}
    \renewcommand{\arraystretch}{0.75}

    \begin{tabular}{@{}lrrrrrr@{}}
        \toprule

        \rowcolor{gray!15}
        \multicolumn{7}{c}{\textbf{LuxEmo speaker-level statistics}} \\
        \midrule

        \rowcolor{gray!05}
        Speaker
        & Segments
        & Duration (min)
        & Neutral
        & Happy
        & Sad
        & Angry \\
        \midrule

        Spk1 (M)
        & 1{,}240
        & 206.7
        & 88
        & 1{,}125
        & 17
        & 10 \\

        Spk2 (F)
        & 766
        & 127.7
        & 190
        & 476
        & 100
        & 0 \\

        Spk3 (F)
        & 3{,}577
        & 596.2
        & 1{,}627
        & 1{,}223
        & 702
        & 25 \\

        Spk4 (F)
        & 1{,}979
        & 329.8
        & 1{,}063
        & 836
        & 77
        & 3 \\

        \midrule

        \textbf{Total (1M, 3F)}
        & \textbf{7{,}562}
        & \textbf{1{,}260.4}
        & \textbf{2{,}968}
        & \textbf{3{,}660}
        & \textbf{896}
        & \textbf{38} \\

        \midrule

        \rowcolor{gray!15}
        \multicolumn{7}{c}{\textbf{General corpus comparison}} \\
        \midrule

        \rowcolor{gray!05}
        Dataset
        & Language
        & Hours
        & Speakers
        & Emotions
        & Style
        & Context \\
        \midrule

        EmoDB
        & de
        & 0.4
        & 10
        & 7
        & acted
        & studio \\

        \textbf{LuxEmo}
        & \textbf{lb}
        & \textbf{21.0}
        & \textbf{4}
        & \textbf{4}
        & \textbf{spontaneous}
        & \textbf{youth media} \\

        \midrule

        \rowcolor{gray!15}
        \multicolumn{7}{c}{\textbf{Linguistic coverage comparison}} \\
        \midrule

        \rowcolor{gray!05}
        Statistic
        & \multicolumn{3}{c}{EmoDB}
        & \multicolumn{3}{c}{LuxEmo} \\
        \cmidrule(lr){2-4}
        \cmidrule(lr){5-7}

        Utterances / segments
        & \multicolumn{3}{c}{535}
        & \multicolumn{3}{c}{7{,}562} \\

        Duration (h)
        & \multicolumn{3}{c}{0.4}
        & \multicolumn{3}{c}{21.0} \\

        Unique sentences
        & \multicolumn{3}{c}{10}
        & \multicolumn{3}{c}{$\sim$7{,}097} \\

        Running words
        & \multicolumn{3}{c}{$\sim$3{,}584}
        & \multicolumn{3}{c}{165{,}587} \\

        Average words / utterance
        & \multicolumn{3}{c}{6.7}
        & \multicolumn{3}{c}{23.3} \\

        Unique words
        & \multicolumn{3}{c}{55}
        & \multicolumn{3}{c}{14{,}520} \\

        Phoneme tokens
        & \multicolumn{3}{c}{15{,}087}
        & \multicolumn{3}{c}{482{,}395} \\

        Diphone tokens
        & \multicolumn{3}{c}{14{,}552}
        & \multicolumn{3}{c}{338{,}446} \\

        Unique diphones
        & \multicolumn{3}{c}{181}
        & \multicolumn{3}{c}{1{,}622} \\

        \bottomrule
    \end{tabular}

    \begin{tablenotes}[flushleft]
        \scriptsize
        \item \textit{Notes.}
    M/F denotes the speaker's gender.
    LuxEmo word counts are derived from LuxASR transcriptions.
    G2P-derived counts use Luxembourgish Sequitur(~\url{https://github.com/PeterGilles/sequitur-g2p}) for LuxEmo and German
    \textit{gruut} for EmoDB;(~\cite{bisani2008joint}
    and available at~\url{https://github.com/sequitur-g2p/sequitur-g2p}).
    \end{tablenotes}

    \end{threeparttable}
\end{table*}

\section{Expressive TTS Systems}
\label{sec:systems}

\subsection{Baseline TTS systems and evaluation protocol}

We evaluate five SOTA expressive TTS systems on LuxEmo, namely GradTTS, XTTS, Toucan, Qwen3\_FT, and kNN TTS. The set covers three zero-shot models and two Luxembourgish-adapted systems. GradTTS, a diffusion-based model, and XTTS, a cross-lingual reference-based model, use German as a close proxy language since neither of them natively supports Luxembourgish. Toucan, a reference encoder-based model, relies on a multilingual phoneme inventory that covers Luxembourgish. Two additional systems are adapted to LuxEmo: a prompt-controlled language model-based TTS and a non-parametric baseline that transfers prosody by retrieving k nearest neighbor segments in LuxEmo~\footnote{\url{https://anonymous.4open.science/r/LuxEmo_Sample-445F/}}. 

Synthesis and evaluation use a reproducible pipeline that standardizes audio generation, metric computation, and figure production. We generate synthetic speech in two configurations: a fixed evaluation set with predefined test sentences per speaker-emotion pair, and a full-corpus pass synthesizing all metadata transcriptions. Metadata for the four RTL speakers is consolidated into an index of 7{,}562 segments with speaker identifier, emotion, transcription, language tag, and gender, enabling stratified sampling and per-group analysis. For the German reference condition, we synthesize $40$ EmoDB sentences~\cite{burkhardt2005database} across four emotions using the diffusion-based model with German settings, leveraging the phonetic overlap between Luxembourgish and standard German as an acted German reference for audio quality and prosody metrics.

We report zero-shot and Luxembourgish-adapted systems separately to clarify their deployment assumptions. The five systems cover practical strategies for Luxembourgish expressive synthesis, including cross-lingual transfer, multilingual support, fine-tuning, and retrieval-based prosody transfer. Since they differ in architecture, training data, language coverage, vocoder, and adaptation procedure, the results should be read as a realistic usability benchmark rather than a controlled architectural ablation. The TTS experiments therefore show that LuxEmo supports training, adaptation, retrieval, and evaluation pipelines, but do not by themselves establish emotion-label accuracy, acoustic cleanliness, or demographic representativeness.

\subsection{Objective and Subjective Evaluation}
Eight objective evaluation metrics are used to evaluate LuxEmo, 
\begin{enumerate*}[label=(\roman*)]
  \item \textbf{Audio quality:} WV-MOS~\cite{andreev2023hifi++} for
        non-intrusive quality prediction
  \item \textbf{TTS naturalness:} NISQA-TTS~\cite{mittag2020deep},
        a non-intrusive deep-learning model predicting the perceived
        naturalness of synthesised speech on a 1--5 MOS scale
  \item \textbf{Multi-dimensional speech quality:} NISQA
        v2.0~\cite{mittag2021nisqa}, predicting overall quality
        (MOS), noisiness, coloration, discontinuity, and loudness
  \item \textbf{Noise-aware quality:} DNSMOS~\cite{reddy2021dnsmos},
        a non-intrusive model providing overall (OVRL), signal (SIG),
        and background-noise (BAK) scores calibrated on noisy and
        enhanced speech; audio is resampled to 16~kHz prior to scoring
  \item \textbf{Generic intelligibility:} WER with Whisper
        base~\cite{radford2023robust}
  \item \textbf{Language-specific intelligibility:} WER and CER with
        LuxASR, a Whisper large-v3 model fine-tuned for Luxembourgish
  \item \textbf{Speaker similarity:} cosine similarity and EER with
        SpeechBrain ECAPA-TDNN embeddings~\cite{ravanelli2021speechbrain}
  \item \textbf{Prosody:} F0 RMSE and Pearson correlation from
        pYIN~\cite{mauch2014pyin} against reference recordings
\end{enumerate*}

For pairwise model comparison, we compute comparative MOS as the mean WV MOS difference over paired samples, and reference-based metrics (speaker similarity and prosody) use original LuxEmo recordings as groundtruth aggregated per model, speaker, and emotion.

We conduct a subjective listening test with stratified sampling by model and target emotion, implemented in Streamlit~\cite{streamlit}. The study involved 20 native Luxembourgish listeners aged 20--50, with balanced gender distribution, self-reported normal hearing, and voluntary participation. Each session lasted 12--15 minutes. Each listener rated $20$ stimuli covering all five TTS systems and four target emotions, one text per system$\times$emotion pair, sampled from the four LuxEmo speakers without enforcing a fully crossed design. For each stimulus, listeners (i) selected the perceived emotion (neutral, happy, sad, angry, other, or unsure), (ii) rated emotion intensity on a $5$-point scale, and (iii) rated how natural and appropriate the emotional tone was for the sentence on a $5$-point scale, yielding $60$ responses per listener and $1{,}200$ responses overall. We report mean scores with $95\%$ confidence intervals, emotion recognition accuracy, and Krippendorff's $\alpha$ for inter-rater agreement on rating scales.

\section{Results and Analysis}
\label{sec:results}
\begin{figure*}[hpt!]
    \centering
    \includegraphics[width=\linewidth]{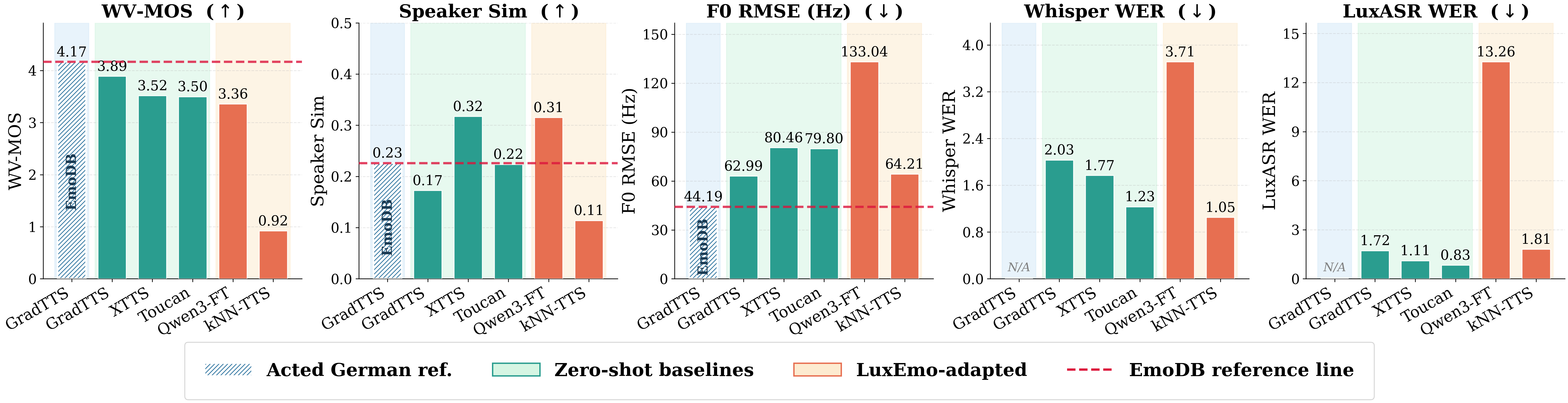}
    \caption{
    Aggregate objective metrics for all evaluated TTS systems. The hatched bar and dashed line mark the acted German reference condition.}
    \label{fig:metrics_all}
\end{figure*}

This section summarizes objective and subjective results. We focus on three questions: overall performance of expressive TTS systems for Luxembourgish, cross-lingual differences between German-proxy and Luxembourgish-aware conditions, and the effect of Luxembourgish-specific adaptation. Figure~\ref{fig:metrics_all} presents objective metrics for all systems, including WV MOS, speaker similarity, F0 RMSE, Whisper WER, and LuxASR WER. Table~\ref{tab:quality_listening} reports complementary non-intrusive quality scores and listening-test results. Together, Figure~\ref{fig:metrics_all} summarizes system-level objective performance, while Table~\ref{tab:quality_listening} provides acoustic-quality estimates and listener judgments.

Figure~\ref{fig:metrics_all} reports the aggregate objective performance of all evaluated systems. The notation \textit{de} refers to German and \textit{lb} to Luxembourgish. GradTTS on EmoDB serves as an acted German reference, with high WV MOS and low F0 RMSE, illustrating the easier conditions of acted, studio-quality speech compared with spontaneous, code-switching Luxembourgish. Among Luxembourgish evaluation conditions, GradTTS~(de) attains the highest WV MOS but also the highest LuxASR WER, indicating that smooth waveforms do not necessarily imply good Luxembourgish segmental accuracy. Toucan~(lb) achieves the lowest LuxASR WER while maintaining competitive quality, and XTTS~(de) yields the highest speaker similarity but only intermediate intelligibility. This pattern is consistent with low-resource TTS work showing that cross-lingual transfer from high-resource languages can improve perceived quality while still leaving gaps in target-language intelligibility~\cite{chen2019end,byambadorj2021text,cai2023cross}. The multi-metric profile in Figure~\ref{fig:radar} makes these trade-offs explicit: no single model dominates across quality, intelligibility, speaker similarity, and prosody.

\subsection{Audio Quality and Noise Assessment}
\label{sec:quality}

Table~\ref{tab:quality_listening} reports non-intrusive quality scores from NISQA-TTS~\cite{mittag2020deep}, NISQA v2.0~\cite{mittag2021nisqa}, and DNSMOS~\cite{reddy2021dnsmos}. The first row covers preprocessed LuxEmo recordings after denoising and VAD-based segmentation, while the remaining rows cover synthesized TTS outputs. Corpus recordings obtain NISQA~=~3.38 and DNSMOS OVRL~=~2.91, providing a source-quality reference for broadcast-origin material. TTS-output scores capture complementary aspects of predicted naturalness, waveform quality, and robustness to residual background noise.

\textbf{Naturalness and waveform quality.}
Toucan~(lb) obtains the highest NISQA-TTS naturalness score (2.95) and NISQA overall quality score (3.94) among synthesized systems. GradTTS~(de) ranks second on NISQA overall quality (3.44), followed by XTTS~(de) (3.20). This suggests Luxembourgish phoneme coverage helps predict naturalness and waveform quality, although German-proxy systems remain competitive.

\textbf{Noise-aware quality.}
XTTS~(de) achieves the highest DNSMOS overall score (2.87) and signal score (SIG~=~3.22), while Toucan~(lb) obtains the highest background score (BAK~=~3.80). GradTTS~(de) and Toucan~(lb) have comparable DNSMOS OVRL scores (2.68 and 2.75), indicating moderate noise-aware quality. XTTS shows stronger signal quality; Toucan shows slightly better background-noise suppression.

\textbf{Adapted systems.} Qwen3\_FT obtains lower NISQA overall quality (1.75) and DNSMOS OVRL (1.95) than zero-shot systems, indicating reduced predicted waveform quality after fine-tuning. kNN TTS obtains higher NISQA overall quality than Qwen3\_FT (2.15 vs. 1.75), but lower DNSMOS scores overall (OVRL~=~1.86, SIG~=~2.52, BAK~=~2.51). The two adapted systems show different weaknesses: Qwen3\_FT is preferred in the listening test despite low predicted quality, while kNN TTS remains limited by acoustic quality.

\textbf{Comparison with WV MOS.}
The metric rankings are not identical. Toucan~(lb) leads on NISQA-TTS and NISQA overall quality, while Figure~\ref{fig:metrics_all} shows a different WV MOS ranking. This divergence shows that objective metrics emphasize different aspects of synthesis quality. WV MOS appears to reward global waveform smoothness, whereas NISQA and DNSMOS penalize different dimensions of synthetic-speech quality, including coloration, discontinuity, signal quality, and background noise. The complementarity of these metrics supports the use of a multi-metric evaluation protocol for expressive low-resource TTS~\cite{mittag2021nisqa,andreev2023hifi++}.

\subsection{Luxembourgish adaptation and cross-lingual effects}
\label{sec:ablation}

To assess Luxembourgish-specific adaptation, we evaluated two LuxEmo-adapted systems: Qwen3\_FT, a per-speaker fine-tuned prompt-controlled TTS model, and kNN TTS, a GlowTTS-based system that transfers prosody through $k$ nearest-neighbor retrieval. Figure~\ref{fig:metrics_all} compares these systems with the zero-shot models.

Qwen3\_FT obtains reasonable WV MOS and high speaker similarity, but its LuxASR WER is much higher than that of the zero-shot systems. This indicates ASR-based segmental mismatch, not direct evidence that listeners found the speech unintelligible. kNN TTS achieves F0 RMSE comparable to GradTTS, showing that it can retrieve LuxEmo prosody, but its low WV MOS indicates weak waveform quality. Overall, Luxembourgish adaptation does not uniformly improve objective metrics. Qwen3\_FT performs well in the listening test but has weak ASR-based intelligibility, while kNN TTS captures prosody but remains limited by acoustic quality.

The comparison between XTTS~(de) and Toucan~(lb) highlights the value of Luxembourgish-aware modeling. On the same evaluation text, Toucan~(lb) improves Whisper WER, LuxASR WER, and F0 RMSE relative to XTTS~(de), despite similar WV MOS and lower speaker similarity. This suggests that German-proxy systems can produce smooth speech but still struggle with Luxembourgish phonotactics and code-switched tokens. Luxembourgish-aware modeling should therefore complement surrogate-language data~\cite{cai2023cross}.

\subsection{Emotion-wise behavior and human judgments}

\begin{figure}[!t]
    \centering
    \includegraphics[width=\columnwidth]{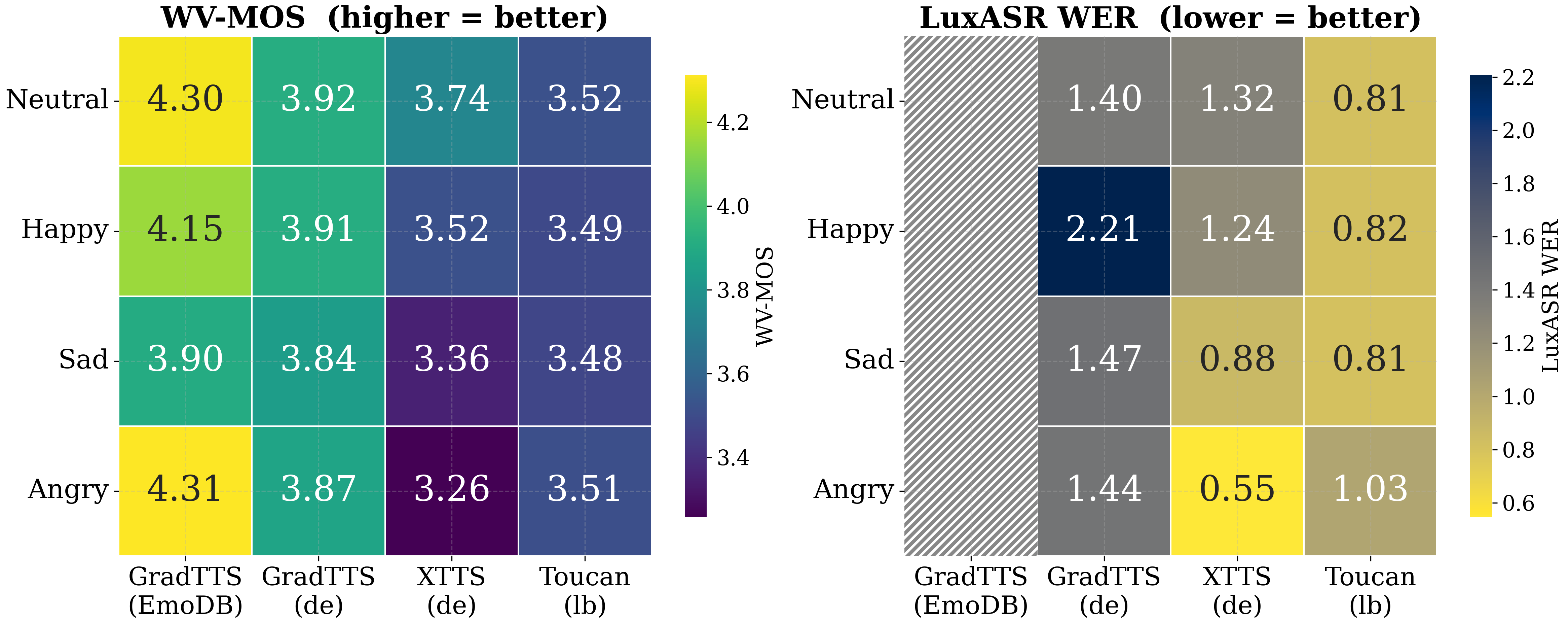}
    \caption{Per-emotion performance breakdown. Left: WV-MOS (higher is better),
    right: LuxASR WER (lower is better).}
    \label{fig:emotion_heatmap}
\end{figure}

Figure~\ref{fig:emotion_heatmap} illustrates WV MOS and LuxASR WER according to emotion and model. GradTTS yields the highest WV MOS across emotions, with angry and neutral scoring best, consistent with acted studio conditions. Among Luxembourgish systems, GradTTS (de) offers perceived quality but high WER, especially for happy, while Toucan (lb) shows stable intelligibility with LuxASR WER in a narrow band across emotions. Angry is consistently hardest for XTTS and Toucan, matching the corpus imbalance where anger covers only $0.5\%$ of segments and limits generalization to high-arousal prosody.

The listening test mirrors the WV MOS results but introduces important nuances. Table~\ref{tab:quality_listening} summarizes the subjective evaluation with 20 native Luxembourgish listeners. Naturalness and emotion appropriateness were rated on five-point scales. The reported values are means with 95\% confidence intervals computed across listener-level means. The listening test did not include a separate subjective intelligibility rating.

\begin{table}[!t]
\centering
\scriptsize
\setlength{\tabcolsep}{1.4pt}
\renewcommand{\arraystretch}{0.7}
\caption{Objective quality and subjective listening-test results.}
\label{tab:quality_listening}
\begin{threeparttable}
\begin{tabular*}{\columnwidth}{@{\extracolsep{\fill}}lrrrrrr@{}}
\toprule
\rowcolor{gray!20}
\multicolumn{7}{c}{\textbf{Objective quality scores}} \\
\midrule
\rowcolor{gray!05}
System & N & \shortstack{NISQA\\TTS} & \shortstack{NISQA\\Overall} & \shortstack{DNSMOS\\Overall} & \shortstack{DNSMOS\\Signal} & \shortstack{DNSMOS\\Background} \\
\midrule
\rowcolor{gray!08}
\multicolumn{7}{@{}l}{\textit{Preprocessed corpus recordings}} \\
LuxEmo orig. & 7{,}562 & -- & 3.38 & 2.91 & 3.34 & 3.73 \\
\midrule
\rowcolor{gray!08}
\multicolumn{7}{@{}l}{\textit{Synthesized TTS outputs}} \\
GradTTS (de) & 612 & 2.85 & 3.44 & 2.68 & 3.05 & 3.72 \\
XTTS (de)    & 612 & 2.81 & 3.20 & \textbf{2.87} & \textbf{3.22} & 3.78 \\
Toucan (lb)  & 612 & \textbf{2.95} & \textbf{3.94} & 2.75 & 3.10 & \textbf{3.80} \\
Qwen3\_FT    & 53  & 2.32 & 1.75 & 1.95 & 2.77 & 2.53 \\
kNN TTS      & 280 & 2.00 & 2.15 & 1.86 & 2.52 & 2.51 \\
\midrule
\rowcolor{gray!20}
\multicolumn{7}{c}{\textbf{Subjective listening-test results}} \\
\midrule
\rowcolor{gray!05}
System & \multicolumn{3}{c}{Naturalness} & \multicolumn{3}{c}{Emotion appropriateness} \\
\cmidrule(lr){2-4} \cmidrule(lr){5-7}
\rowcolor{gray!05}
 & Mean & L & U & Mean & L & U \\
\midrule
Qwen3\_FT    & \textbf{3.9} & \textbf{3.6} & \textbf{4.2} & \textbf{4.0} & \textbf{3.7} & \textbf{4.3} \\
Toucan (lb)  & 3.7 & 3.4 & 4.0 & 3.8 & 3.5 & 4.1 \\
GradTTS (de) & 3.7 & 3.4 & 4.0 & 3.6 & 3.3 & 3.9 \\
XTTS (de)    & 3.4 & 3.1 & 3.7 & 3.4 & 3.1 & 3.7 \\
kNN TTS      & 2.3 & 2.0 & 2.6 & 2.4 & 2.1 & 2.7 \\
\bottomrule
\end{tabular*}
\begin{tablenotes}[flushleft]
\scriptsize
\item \textit{Notes.}
NISQA-TTS is omitted for original recordings. Mean, L, and U denote subjective mean and 95\% CI limits. Bold marks the best synthesized system per metric.

\end{tablenotes}
\end{threeparttable}
\end{table}

Table~\ref{tab:quality_listening} summarizes the listening test with 20 native Luxembourgish listeners. Naturalness and emotion appropriateness were rated on five-point scales, with results reported as means and 95\% confidence intervals computed over listener-level means. The test did not include separate intelligibility ratings.

Qwen3\_FT obtains the highest mean ratings for naturalness and emotion appropriateness, followed by Toucan~(lb) and GradTTS~(de). However, the confidence intervals overlap across the top systems, so this ranking should be interpreted descriptively rather than as a statistically reliable separation. The result is still useful because Qwen3\_FT receives strong listener ratings despite weaker ASR-based intelligibility estimates, showing that perceived expressiveness and ASR-derived transcription accuracy do not always align.

Addressing \textbf{RQ1}, Qwen3\_FT is rated highest by listeners, while Toucan~(lb) and GradTTS~(de) form a second tier. This suggests that Luxembourgish adaptation can improve perceived emotional quality, although the sample size and overlapping confidence intervals limit statistical conclusions. For \textbf{RQ2}, Luxembourgish-aware systems perform better on either emotional appropriateness or ASR-based intelligibility, depending on the system and metric. This indicates that German proxy models do not fully capture Luxembourgish-specific expressive cues. For \textbf{RQ3}, lower LuxASR WER does not consistently predict higher perceived naturalness. ASR-based intelligibility, waveform quality, and listener-rated emotional quality can diverge, so intelligibility alone is not a sufficient proxy for expressive TTS quality.

\subsection{Error patterns and implications}

\begin{figure}[!t]
    \centering
    \includegraphics[width=0.75\linewidth]{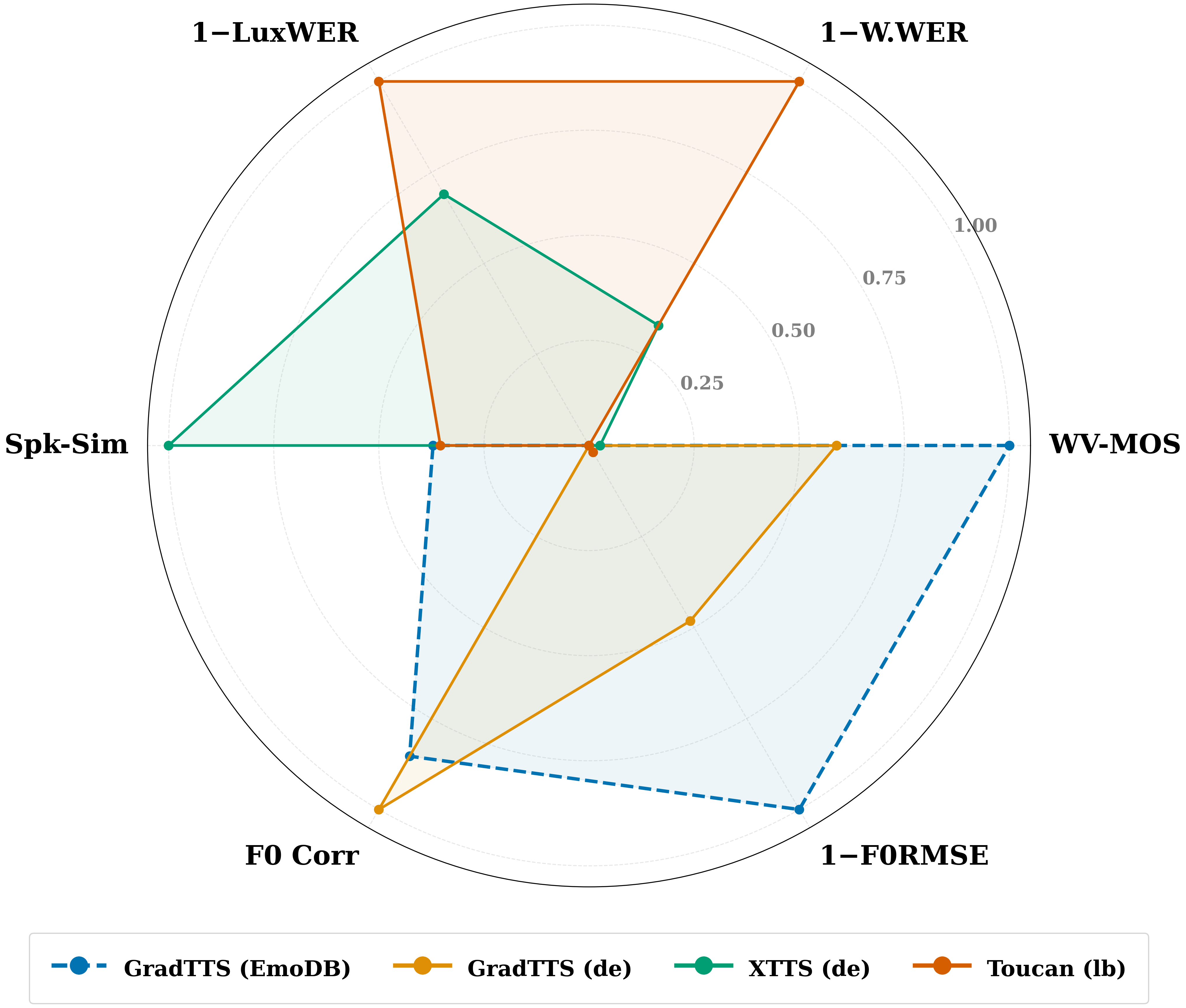}
    \caption{Normalized model profiles (higher is better).}
    \label{fig:radar}
\end{figure}

Qualitative inspection shows that models struggle most on segments with dense code switching, rapid turn-taking, or strong background music in the original LuxEmo recordings, characteristics of spontaneous Luxembourgish speech that are underrepresented in standard emotional TTS corpora, suggesting a mismatch between training data and real-world Luxembourgish usage. Emotion confusions are frequent for subtle negative states, such as low-intensity sadness, and for rare anger segments where training data are scarce. Cross-lingual conditions introduce degradation, including German prosody leaking into Luxembourgish utterances and mispronunciations of Luxembourgish-specific lexical items. This shows that even a closely related proxy language fails to capture the linguistic cues of Luxembourgish, as evident in Figure~\ref{fig:radar}.

Recent emotional TTS work reports that fine-grained control of expressive prosody is fragile under such conditions and requires a richer evaluation than global MOS or WER, for example, combining listening tests with benchmarks targeting emotional and prosodic challenges~\cite{liang2025ece,jing2025enhancing,manku2025emergenttts}.

\subsection{Dataset limitation and scope}
The four-speaker composition is determined by the source material rather than an a priori demographic sampling decision. The selected RTL Youth programs rely on a small set of recurring presenters, while other voices occur sporadically or cannot be identified consistently across episodes. After speaker identification, diarization, and usability filtering, four recurring speakers provided sufficient consistent, attributable speech for corpus construction. Consequently, LuxEmo should not be interpreted as demographically representative of Luxembourgish speakers. Its strength lies in capturing spontaneous, expressive speech from a coherent youth-media domain. 

\section{Conclusion and Future Work} \label{sec:conclusion} We introduced LuxEmo, a spontaneous emotional speech corpus and expressive TTS benchmark for Luxembourgish derived from RTL Youth broadcasts. It contains 21 hours of wideband speech and 7{,}562 ten-second segments from four recurring speakers, annotated for language, speaker identity, and four emotion categories in a code-switching setting. We described a semi-automatic curation workflow that combines VAD, denoising, language identification, LuxASR-based speaker-aware segmentation, acoustic and lexical emotion cues, human review and corpus-quality assessment. We benchmarked five expressive TTS strategies using objective metrics and a native-listener study. The results show that waveform quality, ASR-based intelligibility, prosody, speaker similarity, and emotional appropriateness do not always align. Future work will broaden speaker and domain coverage, refine emotion labels through inter-annotator analysis, and adapt the pipeline to other low-resource, code-switching settings.

\bibliographystyle{IEEEtran}
\bibliography{mybib}
\end{document}